# HAZE-Net: High-Frequency Attentive Super-Resolved Gaze Estimation in Low-Resolution Face Images

Jun-Seok Yun[1], Youngju Na[1], Hee Hyeon Kim[1], Hyung-Il Kim[2], Seok Bong Yoo[1, *]

[1] Department of Artificial Intelligence Convergence, Chonnam National University, Korea
[2] Electronics and Telecommunications Research Institute, Korea

218062@jnu.ac.kr, aperxnt1@gmail.com, muke0822@gmail.com, hikim@etri.re.kr, sbyoo@jnu.ac.kr

**Abstract.** Although gaze estimation methods have been developed with deep learning techniques, there has been no such approach as aim to attain accurate performance in low-resolution face images with a pixel width of 50 pixels or less. To solve a limitation under the challenging low-resolution conditions, we propose a high-frequency attentive super-resolved gaze estimation network, i.e., HAZE-Net. Our network improves the resolution of the input image and enhances the eye features and those boundaries via a proposed super-resolution module based on a high-frequency attention block. In addition, our gaze estimation module utilizes high-frequency components of the eye as well as the global appearance map. We also utilize the structural location information of faces to approximate head pose. The experimental results indicate that the proposed method exhibits robust gaze estimation performance even in low-resolution face images with 28×28 pixels. The source code of this work is available at https://github.com/dbseorms16/HAZE_Net/.

## 1 Introduction

Human gaze information provides principal guidance on a person's attention and is significant in the prediction of human behaviors and speculative intentions. Accordingly, it has been widely used in various applications, such as human-computer interaction [1, 2], autonomous driving [3], gaze target detection [4], and virtual reality [5]. Most of the existing methods for estimating the human gaze by utilizing particular equipment (e.g., eye-tracking glasses and virtual reality/augmented reality devices) are not suitable for real-world applications [6-8]. Recently, to solve this problem, face image's appearance-based gaze estimation methods that learn a direct mapping function from facial appearance or eyes to human gaze are considered. To accurately estimate the human gaze by using these methods, an image with well-preserved eye features (e.g., the shape of the pupil) and well-separated boundaries (e.g., the boundary between the iris and the eyelids) is crucial.
Recent studies [9-10] show reliable performance for the gaze estimation with high-resolution (HR) face images from 448 × 448 to 6000 × 4000 pixels including abundant eye-related features. However, in the real world, even the face region detected from the HR image may have a low-resolution (LR) depending on the distance between the camera and the subject, as shown in the upper-left of Fig. 1. Most of the existing gaze estimation methods use fixed-size images. Thus, when the distance between the camera and the subject is large, it leads to the severe degradation of gaze estimation due to the lack of resolution of the eye patches, as shown in the first row of Fig. 1.

* Corresponding author

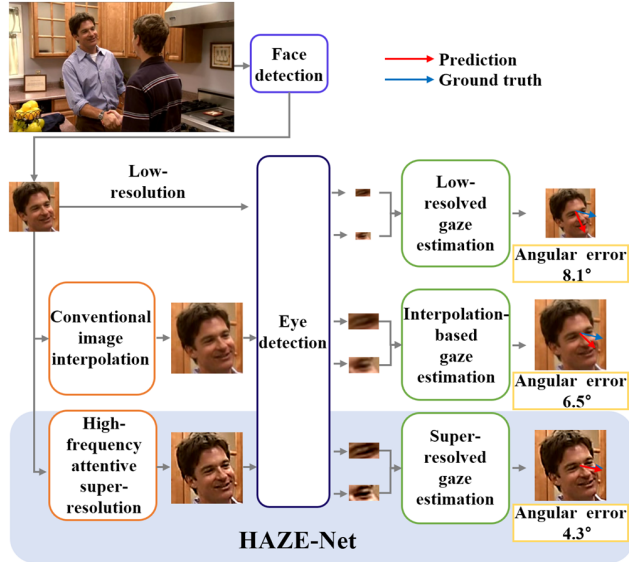

**Fig. 1.** Examples of gaze estimation approaches in the real world. HAZE-Net introduced a high-frequency attentive super-resolved gaze estimation that outperforms conventional methods by a large margin.

To deal with the problem, the conventional image interpolation approach can be adopted so that the image resolution for eye regions can be enhanced as shown in the second row of Fig. 1. However, since it only works based on the limited relationship between the surrounding pixels, this method cannot resolve the degradation of gaze estimation performance. As an alternative, image super-resolution (SR) methods [11-18] have been considered. These methods are performed to restore HR images from the LR images. Accordingly, the SR module learns how to reconstruct the LR images to HR images. However, this is an ill-posed problem with various possible answers. This indicates that conventional SR modules do not stably enhance eye features and boundaries, which are essential for gaze estimation. In other words, although the SR approach may help improve the quality of the image, it does not guarantee an ideal mapping for the optimal performance of the gaze estimation. The results of each image upscaling method for an LR face image with 28×28 pixels are shown in Fig. 2. A severe degradation problem occurs in the up-sampled image when bicubic interpolation is applied to the LR image, as shown in bicubic results. On the other hand, the conventional SR module shows higher quality, as shown in DRN [12] results. Nevertheless, it can be seen that the boundary between the iris and the pupil is not distinctly differentiated. This demonstrates that the conventional SR method does not provide an optimal mapping guideline for gaze estimation.

In this paper, we propose a high-frequency attentive super-resolved gaze estimation network, so-called HAZE-Net, which is mainly comprised of two modules: 1) SR module based on a high-frequency attention block (HFAB) and 2) global-local gaze estimation module. To deal with the limitations in the conventional SR methods, we rei

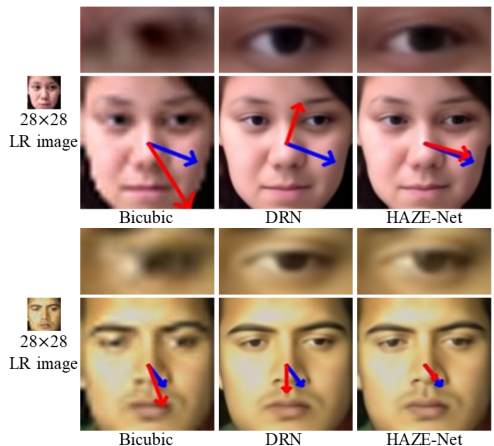

**Fig. 2.** Visual comparison of different 4× up-sampling methods for LR face images with 28 × 28 pixels. The first row for each LR image shows the enlarged eye images using different methods. The second row for each LR image represents the ground truth (blue arrow) and predicted gaze (red arrow), respectively.

-nforce the high-frequency information inspired by the observation that the contour of eye features and their boundaries correspond to high-frequency in the frequency domain. Through the proposed SR module, we observe that it preserves the shape of the pupil well and distinctly differentiates the boundary between the iris and the pupil. In addition to the SR module, we devise a global-local gaze estimation module. Based on the super-resolved face images and corresponding global-local (face-eye) appearance maps are used to improve the gaze estimation performance. In addition, we use the coordinates of five landmarks (e.g., eyes, nose, both corners of the mouth) containing the structural location information of the face to provide an appropriate guide to the head pose. Moreover, the devised two modules are collaboratively trained via the proposed alternative learning strategy. In this process, we add a constraint on the SR module to produce a face image that is favorable to gaze estimation. It contributes to improving gaze estimation performance, as shown in HAZE-Net results in Fig. 2. We test gaze estimation performance under LR conditions using MPIIFaceGaze [19] and EyeDiap [20] datasets. The proposed method effectively estimates the gaze under challenging LR conditions (e.g., 28 × 28 pixels face image). The major contributions of the paper are as follows:

• The HFAB proposed in our SR module strengthens the high-frequency information including eye regions, which is crucial for gaze estimation. With the contribution of HFAB, the LR face image can be enhanced to be suitable for gaze estimation.

• Our gaze estimation module utilizes the global-local appearance map obtained via high-frequency extraction. It improves the performance to be robust to person-specific appearance and illumination changes.

• HAZE-Net performs favorably against typical gaze estimation models under challenging LR conditions.

## 2   Related Work

**Appearance-based Gaze Estimation.** Gaze estimation can be divided into two methods: model-based methods [21-23] and appearance-based methods [24-32]. Model-based methods estimate human gazes from the shape of the pupil and boundaries between the iris and the pupil by handcrafted features. However, recently, appearance-based methods have been in the spotlight owing to large datasets and the advancement of deep learning techniques. These methods learn how to extract embedded features for gaze estimation. As one of the early-stage methods, GazeNet [24] takes a grayscale eye patch as input and estimates the gaze vector. It shows a better gaze estimation performance by additionally using a head pose vector. As an extended version, the performance was further improved by using the VGG network [33]. Spatial-Weights CNN [25] utilizes not only the eye region but also full-face images. Spatial weights are used to encode important positions in the face image. This method weights regions of the face that are useful for gaze estimation. Through this, more weight is assigned to a specific area in the face image. Furthermore, iTracker [26] receives two eyes, face, and face positions as input and predicts the gaze vector. Dilated-Net [27] utilizes dilated-convolutions to extract high-level features without reducing spatial resolution. It is to capture such small changes in eye appearance. Kang Wang et al. [28] point out the difficulty of generalizing gaze estimation because of appearance variations, head pose variations, and over-fitting issues with point estimation. To deal with these issues, they introduced adversarial learning and Bayesian framework in their network so that it can be practically used in real-world applications. Focusing on the fact that over-parameterized neural networks can be quickly over-fitted, disentangling transforming encoder-decoder (DT-ED) [29] performs few-shot adaptive gaze estimation for learning person-specific gaze networks with very few calibration samples. Coarse-to-fine adaptive network (CA-Net) [30] extracts coarse-grained features from face images to estimate basic gaze direction and fine-grained features from eye images to estimate gaze. In addition, according to the above strategy, they design a bigram model that connects two gaze directions and a framework that introducing attention components to adaptively acquire appropriate subdivided functions. Additionally, there is an attempt to support gaze estimation by utilizing semantic segmentation which identifies different regions of the eyes such as the pupils and iris pupils. RITnet [31] exploits boundary-aware loss functions with a loss scheduling strategy to distinguish coherent regions with crisp region boundaries. PureGaze [32] purifies unnecessary features for gaze estimation (e.g., illumination, personal appearance, and facial expression) through adversarial training with a gaze estimation network and a reconstruction network. Nevertheless, the appearance-based gaze estimation can have a high variance in performance depending on the person-specific appearance (e.g., colors of pupil and skin). In this paper, we devise the global-local appearance map for the gaze estimation to be robust to person-specific appearance. Also, our gaze estimation module effectively learns high-frequency features to be robust to illumination and resolution changes.

**Unconstrained Gaze Estimation.** Despite the emergence of appearance-based methods for gaze estimation, there are limitations on estimating gaze from real-world images owing to various head poses, occlusion, illumination changes, and challenging LR conditions. According to the wide range of head pose, obtaining both eyes in the occluded or illuminated image is difficult. Park et al. [10] proposed a model specifically designed for the task of gaze estimation from single eye input. FARE-Net [9] is inspired by a condition where the two eyes of the same person appear asymmetric because of illumination. It optimizes the gaze estimation results by considering the difference between the left and right eyes. The model consists of FAR-Net and E-net. FAR-Net predicts 3D gaze directions for both eyes and is trained with an asymmetric mechanism. The asymmetric mechanism is to sum the loss generated by the asymmetric weight and the gaze direction of both eyes. E-Net learns the reliability of both eyes to balance symmetrical and asymmetrical mechanisms. To solve this problem in a different approach, region selection network (RSN) [33] learns to select regions for effective gaze estimation. RSN utilizes GAZE-Net as an evaluator to train the selection network. To effectively train and evaluate the above methods, unconstrained datasets which are collected in real-world settings have been emerged [34-38]. Recently, some studies have introduced self-supervised or unsupervised learning to solve the problem of the lack of quantitative real-world datasets [39-41]. In addition, GAN-aided methods [45-46] can be applied to solve the lack of datasets problem. The above studies have been conducted to solve various constraints, but studies in unconstrained resolutions are insufficient. When the recently proposed gaze estimation [24, 27, 32, 42] modules are applied to the LR environment, it is experimentally shown that the performance of these modules is not satisfactory. To deal with this, we propose HAZE-Net which shows an acceptable gaze accuracy under challenging LR conditions.

## 3 Method

This section describes the architecture of the proposed high-frequency attentive super-resolved gaze estimation, that so-called HAZE-Net. The first module for the proposed method is the SR module based on HFABs that is a key component to strengthen the high-frequency component of LR face images. The second module is the global-local gaze estimation module, where discriminative eye features are learned. Note that two modules are collaboratively learned. The overall architecture of HAZE-Net is shown in Fig. 5.

### 3.1 Super-Resolution Module

Our SR module is mainly composed of HFABs to exaggerate the high-frequency components that are highly related to gaze estimation performance. Fig. 3 shows the high-frequency extractor (HF extractor) for extracting high-frequency components from the input. We use the DCT principle that indicates that the more directed from the top-left to the bottom-right in the zigzag direction, the higher is the frequency component. 2D-DCT denoted by $\mathcal{F}$ transforms input $I$ into the DCT spectral domain $D$ for each channel:

$$\boldsymbol{D}^{(i)} = \mathcal{F}\big(\boldsymbol{I}^{(i)}\big), \ \ i = 1,2,\dots n, \qquad (1)$$

where $i$ is a channel index, and $n$ is the number of channels. We create a binary mask $\mathcal{M}$ by using a hyper-parameter λ which decides the masking point as follows:

$$\mathcal{M}_{(x,y)} = \begin{cases} 0, & y < -x + 2\lambda h \\ 1, & otherwise \end{cases}, \qquad (2)$$

where $h$ denotes the height of $I$, and $x$, $y$ denote the horizontal and vertical coordinates of $\mathcal{M}$, respectively. The size of $\mathcal{M}$ equals $I$. The hyper-parameter $\lambda$ ranges from 0 to 1. If the $\lambda$ is too small, overfitting occurs because finer features with low-frequency are emphasized and used for learning. On the other hand, if the $\lambda$ is too large, most of the useful information for gaze estimation such as the shape of the pupil and the boundaries between the iris and the eyelids is lost, preventing performance improvement.

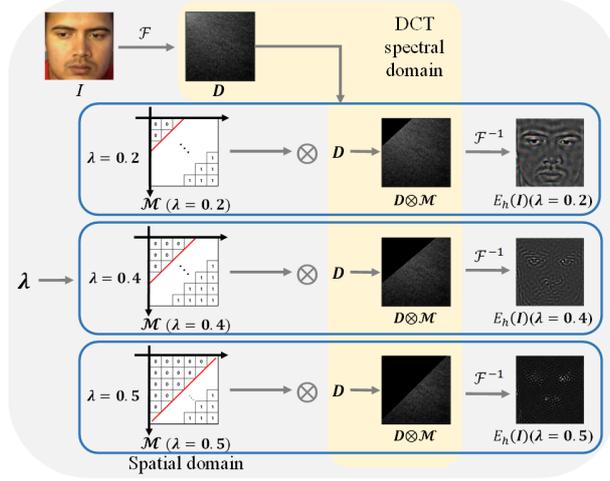

**Fig. 3.** Architecture of HF extractor. Given the spatial domain image or feature map, we map the spatial domain into DCT spectral domain. $\lambda$ is a hyper-parameter that indexes the high-frequency component to be extracted from the top-left to the bottom-right in the zigzag direction. The mask determined by $\lambda$ is multiplied by the feature of the DCT spectral domain. We finally get the high-frequency spatial domain image through 2D-IDCT.

The high-frequency can be separated by element-wise product of $D$ and $\mathcal{M}$. The high-frequency features in the DCT spectral domain are transformed into the spatial domain through 2D-IDCT denoted by $\mathcal{F}^{-1}$:

$$E_h(I) = \mathcal{F}^{-1}(D \otimes \mathcal{M}), \qquad (3)$$

where $\otimes$ denotes the element-wise product and $E_h$ denotes a HF extractor.

HFAB utilizes the residual channel attention block (RCAB) [14] structure, as shown in Fig. 4. RCAB extracts informative feature components to learn the channel statistic. The high-frequency feature map extracted by the HF extractor and the original feature map are assigned to the HFAB as input, as shown in Fig. 4. The original feature map is reinforced by a residual group stacked RCABs for image restoration. To exaggerate the insufficient high-frequency in the original process, the high-frequency feature map is input through a module consisting of two RCABs. Two enhanced results are added to obtain a high-frequency exaggerated feature map. This result becomes a feature in which the outline of the face and the boundary between the elements of the eyes are emphasized.

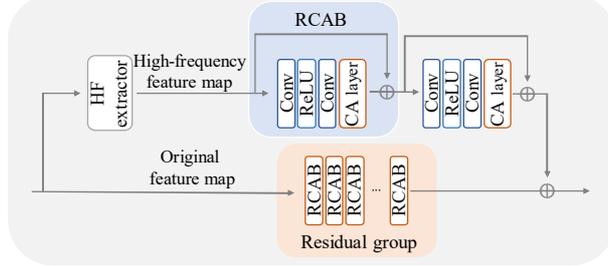

**Fig. 4.** Architecture of HFAB. Given a feature map, we separate the image into the high-frequency feature map and the original feature map. Both features are fed through independent RCABs. We empirically employ two RCABs for the high-frequency feature map and five RCABs for the original feature map. CA layer allows RCAB to learn useful channel-wise features and enhance discriminative learning ability.

Our SR module is composed of the HFABs, $\mathcal{H}$, and the HF extractors, $E_h$. The architecture of our SR module is given on the left side of Fig 5. The module takes an LR image as input and magnifies it to a target size through bicubic interpolation. After the head layer extracts the features from the magnified input, the feature size is reduced through the down block consisting of two convolution layers and a LeakyReLU activation layer. The original features and high-frequency features extracted by $E_h$ are passed through the group of HFABs. The high-frequency enhanced feature through the HFAB is upscaled to the target size through up-block, $U_\uparrow$, consisting of a pixel shuffle layer and two convolution layers. Finally, this extended feature is concatenated with the feature extracted from the head layer and converted into the super-resolved RGB image $I_{SR}$ through the tail layer as follows:

$$I_{SR} = U_\uparrow\big(\mathcal{H}(E_h(f_d), f_d)\big) \oplus f_b, \tag{4}$$

where $f_b$ is the feature extracted from a bicubic-upsampled image, $f_d$ is the feature reduced in size by the downblock and $\oplus$ is the concatenation operation.

### 3.2 Gaze estimation module

The performance of the appearance-based gaze estimation module depends on the resolution of the image received as an input. In general, the proportion of a face in the image is usually small and variable. Thus, resizing the LR face image to a larger size causes severe loss of information that is important to gaze estimation. Therefore, in this paper, we propose the super-resolved gaze estimation module that is robust under LR conditions. As our gaze estimation module adaptively learns through super-resolved images with exaggerated high-frequency, it preserves information that helps estimate gaze under the LR environment. Our module secures stable input by adding additional high-frequency components that are insensitive to this environment. The gaze estimation module has a high variance in performance depending on the appearance of a person. This is because face images contain redundant low-frequency information.

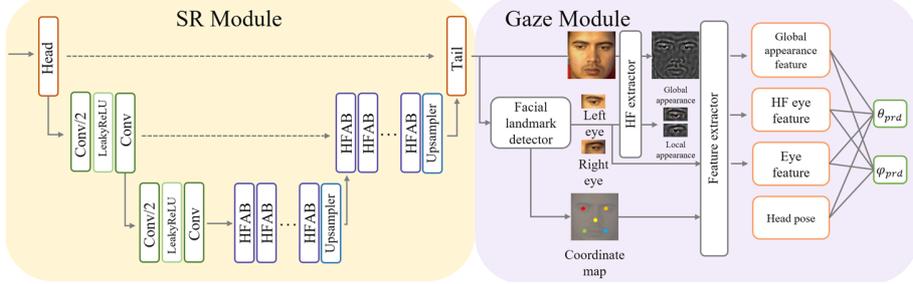

**Fig. 5.** HAZE-Net architecture. In the first module (yellow panel), given the face image of an LR image, we employ our SR module based on the HFAB. The input image goes through one or two down samples according to the scale factor. In the second module (violet panel), our gaze estimation module utilizes eye patches and the global-local appearance map. We feed four features to the final fully connected layer to obtain the estimated gaze angle ($\theta_{prd}, \varphi_{prd}$).

Thus, unnecessary information should be excluded while high-frequencies that help gaze estimation remain. We improve our generalization ability by obtaining a high-frequency appearance map through an HF extractor and using it as an input of the feature extractor. Additionally, we utilize five landmark coordinates such as eyes, nose, and corners of the mouth in the input image during the training process, with the facial landmark detector [43]. The above five coordinates are the structural location information of the face that can be used as a proper guidance of head pose. Our gaze estimation module is designed to receive a super-resolved image and generates five inputs consisting of a high-frequency global appearance map, two high-frequency local maps for each eye, and two eye images. The global appearance map refers to the features that only leave high-frequency features extracted from the face and facial landmarks. Meanwhile, the local appearance map is only extracted from eye patches by the same procedure. It utilizes Resnet-18 as the backbone to extract features from each input. The five features extracted from each input are concatenated into a vector and put into a fully connected layer of size 512. A two-dimensional head-pose vector is used to train our gaze estimation module that predicts a gaze angle ($\theta_{prd}, \varphi_{prd}$).

**3.3 Loss Function**

HAZE-Net employs two loss functions for the SR and gaze estimation modules. The two loss functions are appropriately combined according to the proposed alternative learning strategy.

**SR Loss.** Our module uses L1 Loss, which is commonly used in SR tasks. The loss function minimizes the difference in pixel values between the original image and the SR result image as follows:

$$L_{SR} = \frac{1}{N} \| F_{SR}(\boldsymbol{I}_{LR}) - \boldsymbol{I}_{HR} \|_1, \qquad (5)$$

where $N$ is the image batch size, $\boldsymbol{I}_{LR}$ is the LR image taken as the input, and $\boldsymbol{I}_{HR}$ is the original HR image. In addition, $F_{SR}$ is our high-frequency attention SR module.

**Gaze Estimation Loss**. The proposed gaze estimation module predicts $(\theta, \varphi)$ by projecting a three-dimensional vector $(x, y, z)$ into two dimensions. The predicted ($\theta_{prd}, \varphi_{prd}$) are compared with the ground truth ($\theta_{gt}, \varphi_{gt}$) and the mean squared error as the loss function. Gaze estimation loss is defined as follows:

$$L_{GE} = \frac{1}{N}\sum_{i=1}^{n}((\theta_{prd^i} - \theta_{gt^i})^2 + (\varphi_{prd^i} - \varphi_{gt^i})^2). \qquad (6)$$

**Total Loss.** The total loss function is a combination of the SR loss and the gaze estimation loss. Therefore, the total loss is defined as follows:

$$L_{Total} = L_{SR} + \alpha L_{GE}, \qquad (7)$$

where $\alpha$ is a hyper-parameter that scales gaze estimation loss. If the loss scale is focused on one side, it tends to diverge. Thus, it should be appropriately tuned according to the purpose of each phase. The detailed hyper-parameters according to the phase are introduced in section 3.4.

### 3.4 Alternative End-to-End Learning

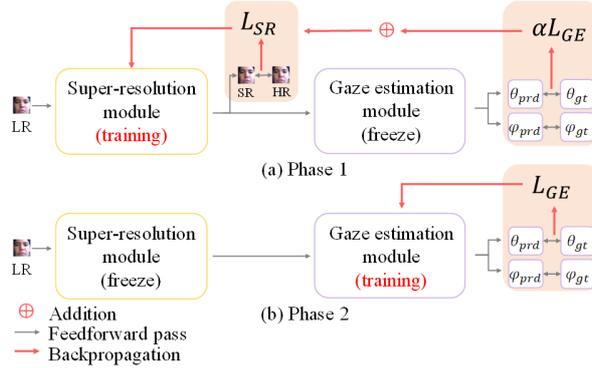

**Fig. 6.** Flowchart of HAZE-Net's alternative end-to-end learning architecture. (a) Phase 1: SR module training while freezing gaze estimation module. (b) Phase 2: Gaze estimation module training while freezing SR module.

This section describes the learning strategies for the proposed HAZE-Net. It is not simply a structural combination of the SR module and the gaze estimation module but a complementary combination through the proposed alternative end-to-end learning. We initialize each module with pre-trained parameters for each task. To train the end-to-end model stably, we combine two modules and apply different losses at each phase, as shown in Fig. 6. We found that using training on our module is more efficient and effective than training from scratch. In phase 1, the SR module is trained while the gaze estimation module is frozen. We use the weighted sum of the SR loss and gaze estimation loss, $L_{Total}$, as shown in Fig. 6(a). We combine two loss functions for our modules to learn complementarily. α is used to perform training by scaling the magnitude of SR loss and gaze estimation loss. Given the scale difference between the two losses, we found that setting the α to 0.1 is the best empirical choice.

In phase 2, the gaze estimation module is trained while the SR module is frozen, as shown in Fig. 6(b). We use only gaze loss $L_{GE}$ in phase 2. Although training both modules without freezing is a possible option, we found that the performance was poor compared with our strategy. The SR images produced by our end-to-end trained SR module generally show clear boundaries between the components of the eyes and the clear shape of the pupil. Although our module may not guarantee a better peak signal-to-noise ratio (PSNR) score, it performs better in the gaze estimation task than simply combining the two separated state-of-the-art (SOTA) modules.

## 4  Experiments

**4.1 Datasets and Evaluation Metrics**

**Datasets.** Based on the datasets used in recent studies [9, 29, 42], we accordingly evaluate our module on the MPIIFaceGaze [19] and EyeDiap [20] datasets. To simulate LR conditions, we set the HR size to 112 × 112 and set LR size according to the scale factor (e.g., 56 × 56, 37 × 37, 28 × 28). For example, if the scale factor is 2×, the resolution of the LR image is 56×56. If the scale factor is 4×, the resolution of the LR image is 28×28. The MPIIFaceGaze contains a total of 45,000 images from 15 subjects. We split 9,000 images of three subjects (p00, p02, p14) for validation, and others are used for the training set. The EyeDiap contains a total of 94 video clips from 16 subjects. We prepare 16,665 images as in [30]. We split 2,384 images consisting of two subjects (p15, p16) and others are utilized for the training set. When generating LR images, we utilize the built-in resize function of MATLAB.

**Environment.** Our module is implemented in PyTorch 1.8.0, and experiments for comparison with other modules are conducted in the same environment. We train each module for 100 epochs with a batch size of 80 and use the Adam optimizer. In addition, we empirically set the hyper-parameter $\lambda$ to 0.2

**Evaluation Metric.** We compare the proposed SR module with SOTA SR methods. For qualitative comparison, we compared the PSNR and structural similarity index measure (SSIM) [44] values of the different methods for scale factors (2×, 3×, and 4×). Also, we compare the proposed gaze estimation module with other gaze estimation modules. We compute the angular error between the predicted gaze vector and the ground-truth gaze vector, and represent the performance of the module as an angular error to numerically show the performance.

**4.2. Performance Comparison by Module**

**Comparison of SR Modules.** We compare our SR module with SOTA SR modules [11-13] in terms of both quantitative results and visual results. All SR modules are trained according to their losses and training methods on MPIIFaceGaze and Eyediap datasets from scratch. Note that gaze datasets are rescaled to simulate low-resolution constraint settings. Therefore, the SR losses are calculated between the SR result and HR image. We present a comparison in terms of high-frequency restoration. As shown in Fig 7, the proposed module enhances the lines, which are high-frequency components, better than DBPN [11] and DRN [12]. SwinIR [13] is comparable to our SR module. As shown in Table 1, the HAZE- Net shows a lower tendency in terms of PSNR and SSIM than the SOTA SR modules. However, as shown in Fig. 7, the proposed HAZE-Net can adequately enhance high-frequency components to be suitable for gaze estimation task that requires clear boundaries. To prove the superiority of our SR module, we measure angular errors on each SR result with the baseline gaze estimation module consisting of ResNet-18 and fully connected layers. As shown in Table 1, our SR module provides the lowest angular error compared with other SOTA SR modules. The HFAB proposed in our SR module strengthens the high-frequency information such as eye features (e.g., the shape of the pupil) and boundaries (e.g., the boundary between the iris and the eyelids). It leads improvement of gaze estimation performance. Moreover, our SR module can restore clean HR image robust to noise of the input image even while maintaining the high-frequency information.

Table 1. Performance comparison with SOTA SR modules for 2×, 3×, and 4×. The best and the second-best results are highlighted in red and blue colors, respectively.

| SR module | Scale | MPIIFaceGaze | | EyeDiap | |
|---|---|---|---|---|---|
| | | PSNR/SSIM | Angular error | PSNR/SSIM | Angular error |
| Bicubic | 2 | 30.83/0.8367 | 7.23 | 35.62/0.9436 | 5.96 |
| DBPN [11] | | 34.35/0.8882 | 6.64 | 39.61/0.9716 | 5.64 |
| DRN [12] | | 33.73/0.8128 | 6.46 | 38.70/0.9228 | 5.44 |
| SwinIR [13] | | 34.40/0.8911 | 6.51 | 40.36/0.9735 | 6.47 |
| Our SR module | | 34.28/0.8263 | 6.23 | 39.65/0.9041 | 4.68 |
| Bicubic | 3 | 26.23/0.6939 | 7.73 | 31.46/0.8722 | 5.64 |
| DBPN [11] | | 31.43/0.8257 | 6.69 | 37.02/0.9447 | 5.18 |
| DRN [12] | | 31.59/0.8279 | 8.52 | 36.19/0.9165 | 6.55 |
| SwinIR [13] | | 31.67/0.9086 | 6.62 | 36.93/0.9657 | 5.32 |
| Our SR module | | 31.33/0.8219 | 6.49 | 36.82/0.9392 | 4.96 |
| Bicubic | 4 | 25.84/0.6429 | 9.32 | 29.58/0.8066 | 6.22 |
| DBPN [11] | | 29.69/0.7704 | 7.06 | 34.96/0.9128 | 5.83 |
| DRN [12] | | 29.77/0.7735 | 6.85 | 33.42/0.8516 | 5.82 |
| SwinIR [13] | | 30.26/0.8723 | 7.54 | 34.02/0.9452 | 5.79 |
| Our SR module | | 29.59/0.7769 | 6.60 | 32.73/0.8934 | 5.54 |

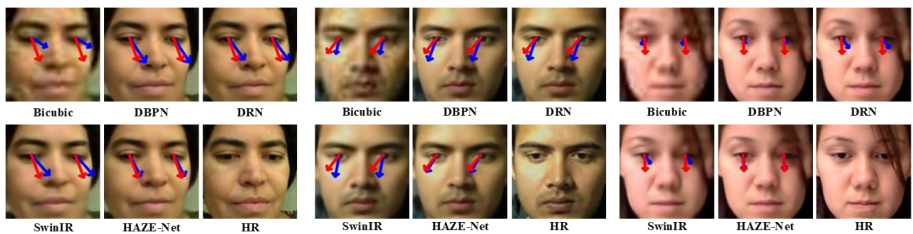

**Fig. 7.** Qualitative comparison of our SR module with SOTA modules on 4× SR.

**Comparison of Gaze Estimation Module.** In this section, we compare the results of our gaze estimation module with other gaze estimation modules. We select a gaze estimation module to compare with our module. Among the recent gaze estimation modules, we exclude modules that use few data [39], unlabeled [40, 41]. All gaze estimation module is trained using an image of size 112 × 112 from scratch. For a fair comparison, we commonly use ResNet-18 as a backbone of all gaze estimation modules. As presented in Table 2, our method shows the best gaze angular error on the MPIIFace-Gaze dataset and the second-best gaze angular error on the EyeDiap dataset. It indicates the superior ty of our gaze estimation module due to the global-local appearance map. In particular, as shown in Fig. 8, our module shows robust performance under challenging illumination conditions.

Table 2. Performance comparison with gaze estimation modules for 112 × 112 HR images. The best and the second-best results are highlighted in red and blue colors, respectively.

| Gaze estimation module | MPIIFaceGaze angular error | EyeDiap angular error |
|---|---|---|
| GazeNet [24] | 5.88 | 4.25 |
| RT-GENE [42] | 5.52 | 4.65 |
| DilatedNet [27] | 5.03 | 4.53 |
| PureGaze [32] | 5.71 | 3.88 |
| Our gaze estimation module | 4.95 | 4.12 |

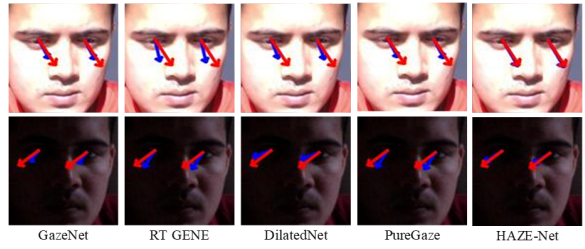

Fig. 8. Visualization of gaze estimation results under challenging illumination conditions. Blue and red arrows represent ground truth and predicted gaze, respectively.

### 4. 3 Comparison under LR Conditions

Table 3. Performance comparison with SOTA SR modules combined with gaze estimation modules under LR conditions. The best and the second-best results are highlighted in red and blue colors, respectively.

| SR module | Gaze estimation module | MPII FaceGaze angular error | EyeDiap angular error | SR module | Gaze estimation module | MPII FaceGaze angular error | EyeDiap angular error |
|---|---|---|---|---|---|---|---|
| LR (56×56) | GazeNet | 9.42 | 9.17 | LR (37×37) | GazeNet | 9.82 | 9.87 |
|  | RT-GENE | 10.13 | 10.33 |  | RT-GENE | 10.54 | 11.41 |
|  | DilatedNet | 15.47 | 17.21 |  | DilatedNet | 16.59 | 18.23 |
| Bicubic 2× (112×112) | GazeNet | 6.89 | 4.46 | Bicubic 3× (111×111) | GazeNet | 8.00 | 5.02 |
|  | RT-GENE | 6.23 | 4.96 |  | RT-GENE | 7.33 | 5.64 |
|  | DilatedNet | 5.59 | 4.55 |  | DilatedNet | 6.85 | 5.93 |
|  | PureGaze | 6.71 | 4.25 |  | PureGaze | 7.92 | 4.91 |
| DBPN 2× (112×112) | GazeNet | 6.07 | 4.05 | DBPN 3× (111×111) | GazeNet | 6.55 | 4.15 |
|  | RT-GENE | 6.59 | 4.82 |  | RT-GENE | 5.65 | 4.79 |
|  | DilatedNet | 5.21 | 4.58 |  | DilatedNet | 5.60 | 5.22 |
|  | PureGaze | 5.94 | 3.98 |  | PureGaze | 6.38 | 3.92 |
| DRN 2× (112×112) | GazeNet | 6.17 | 4.36 | DRN 3× (111×111) | GazeNet | 6.59 | 4.52 |
|  | RT-GENE | 5.76 | 4.69 |  | RT-GENE | 6.52 | 5.38 |
|  | DilatedNet | 5.14 | 5.04 |  | DilatedNet | 5.73 | 5.18 |
|  | PureGaze | 6.01 | 5.51 |  | PureGaze | 6.21 | 5.72 |
| SwinIR 2× (112×112) | GazeNet | 6.47 | 4.21 | SwinIR 3× (111×111) | GazeNet | 7.25 | 4.51 |
|  | RT-GENE | 5.54 | 4.76 |  | RT-GENE | 6.46 | 5.02 |
|  | DilatedNet | 5.03 | 4.39 |  | DilatedNet | 5.56 | 4.47 |
|  | PureGaze | 5.77 | 4.33 |  | PureGaze | 7.09 | 4.39 |
| HAZE-Net 2× (112×112) |  | 4.93 | 3.90 | HAZE-Net 3× (111×111) |  | 5.14 | 3.74 |

| SR module | Gaze estimation module | MPIIFaceGaze angular error | EyeDiap angular error |
|---|---|---|---|
| LR (28×28) | GazeNet | 10.45 | 11.53 |
| | RT-GENE | 10.76 | 12.69 |
| | DilatedNet | 17.89 | 19.23 |
| Bicubic 4× (112×112) | GazeNet | 9.23 | 5.57 |
| | RT-GENE | 9.32 | 6.22 |
| | DilatedNet | 7.52 | 6.14 |
| | PureGaze | 9.17 | 4.96 |
| DBPN 4× (112×112) | GazeNet | 7.10 | 4.81 |
| | RT-GENE | 7.03 | 5.70 |
| | DilatedNet | 5.89 | 4.51 |
| | PureGaze | 6.94 | 4.45 |
| DRN 4× (112×112) | GazeNet | 7.04 | 4.90 |
| | RT-GENE | 7.05 | 5.65 |
| | DilatedNet | 5.82 | 5.39 |
| | PureGaze | 6.88 | 4.30 |
| SwinIR 4× (112×112) | GazeNet | 8.14 | 5.21 |
| | RT-GENE | 7.44 | 5.50 |
| | DilatedNet | 6.38 | 4.67 |
| | PureGaze | 7.97 | 4.56 |
| HAZE-Net 4× (112×112) | | 5.56 | 4.02 |

To verify performance under the LR conditions, we compare HAZE-net, and the combination of SR module and gaze estimation module. Each module is trained with gaze datasets accordingly Tables 1 and 2. For fair comparison under the LR conditions, each gaze estimation modules are fine-tuned with the results of SR modules. Moreover, we set the gaze estimation baselines that are trained with LR image.

In section 4.2, the results show that our module presents lower PSNR and SSIM than those of SwinIR. In contrast, HAZE-Net exhibits the lowest angular error, as presented in Table 3. This is because HAZE-Net successfully enhances high-frequency components, which are critical for gaze estimation performance, compared to other SR modules.

**4.4 Ablation Study**

**Effect of HF Extractor.** We investigate the impact of the HF extractor in order to verify the element performance of our first contribution introduced in section 1. We measure the angular error by using our gaze estimation module trained with HR images. As presented in Table 4, although the HF extractor shows a lower PSNR and SSIM, it exhibits better gaze estimation performance. This indicates that the proposed module enhanced high-frequency components that are suitable for gaze estimation tasks.

Table 4. Quantitative results for evaluating the effects of the HF extractor on the MPIIFaceGaze dataset. The experiment is conducted for 4×.

| HF extractor | PSNR | SSIM | Angular error |
|---|---|---|---|
| ✗ | 30.80 | 0.8397 | 7.27 |
| ✓ | 29.59 | 0.7769 | 6.60 |

**Global-Local Appearance Map.** In this section, we demonstrate the effectiveness of the global-local appearance map introduced as our second contribution. Table 5 shows that both global and local appearance maps help to improve gaze estimation performance. In particular, using the global-local appearance map provides 0.61° lower angular error than using only RGB eye images.

Table 5. Quantitative results for evaluating the effects of global and local appearance maps. The experiments are conducted under HR (112 × 112) conditions.

| Global map | Local map | RGB eye patch | MPIIFaceGaze angular error |
|---|---|---|---|
| ✗ | ✗ | ✓ | 5.56 |
| ✗ | ✓ | ✓ | 5.27 |
| ✓ | ✗ | ✓ | 5.23 |
| ✓ | ✓ | ✓ | 4.95 |

**Hyper-parameters.** We clarify and specify how we decided the value of the hyper-parameters $\alpha$, and $\lambda$. As present in Table 6, when $\alpha$ is 0.1, $\lambda$ is 0.2, it provides the best performance in terms of gaze estimation. $\alpha = 0.1$ means effectiveness of end-to-end learning. $\lambda = 0.2$ is used to improve the generalization performance of HAZE-Net. If the $\lambda$, is too large (e.g., 0.4,0.5), most of the useful information for gaze estimation is lost. We determined the hyper-parameters according to these results.

Table 6. Gaze estimation performance of different levels of $\alpha$ and $\lambda$. The best results are highlighted in red.

| MPIIFaceGaze angular error (◦) | | | |
|---|---|---|---|
| $\alpha$ | 0 | 0.1 | 1 |
| HAZE-Net 4× | 5.47 | 4.95 | 5.13 |
| $\lambda$ | 0.2 | 0.4 | 0.5 |
| HAZE-Net 4× | 4.95 | 5.19 | 5.71 |

**Limitations.** Our module may still be somewhat limited in two aspects in its practical application. In the first aspect, the inference time of our module in 112 × 112 HR resolution that is measured in 2× is 46ms, 3× is 41ms, and 4× is 103ms with NVIDIA RTX 3080 GPU. Therefore, it is slightly difficult to apply in an environment that requires real-time. Second, as our experiment assumes only the bicubic kernel, there is a possibility that the performance will deteriorate in a real environment where the blur kernel is blinded.

## 5. Conclusion

In this paper, we propose a high-frequency attentive super-resolved gaze estimation network. In the SR module, we introduce the HFAB to effectively exaggerate high-frequency components for gaze estimation. In the gaze estimation module, we introduce the global-local high-frequency appearance map. Furthermore, alternative end-to-end learning is performed to effectively train our module. With the contribution of techniques described above, HAZE-Net significantly improves the performance of the gaze estimation module under LR conditions. Extensive experiments including ablation studies demonstrate the superiority of our method over the existing methods.